\documentclass[runningheads]{llncs}
\usepackage[T1]{fontenc}
\usepackage{graphicx}
\usepackage{xcolor}
\usepackage{amsmath,amssymb,amsfonts}
\usepackage{bm}
\usepackage{bbm}
\usepackage{csquotes}
\usepackage{enumitem}
\usepackage{hyperref}
\usepackage{booktabs}
\usepackage{cleveref}
\usepackage{siunitx}

\usepackage[nolist]{acronym}
\begin{acronym}
    \acro{gp}[GP]{Gaussian Process}
    \acro{lmc}[LMC]{linear model of coregionalization}
    \acro{nll}[NLL]{negative log-likelihood}
    \acro{psd}[PSD]{positive-semi-definite}
    \acro{mse}[MSE]{mean squared error}

    \acro{mvts}[MVTS]{multivariate time series}
    \acro{ml}[ML]{machine learning}
    \acro{tmp}[tmp]{tmp}
\end{acronym}

\usepackage{comment}

\usepackage{latexsym}
\usepackage{cite}            
\usepackage{multicol}        
\usepackage{pdfsync}            
\usepackage[active]{srcltx} 
\usepackage{fixmath}        

\crefname{property}{property}{properties}
\Crefname{property}{Property}{Properties}

\DeclareMathOperator*{\argmin}{arg\,min}

\begin{document}
\title{Generative Modeling of Approximately Periodic Time Series by a Posterior-Weighted Gaussian Process}
\titlerunning{PWGP}
%
\author{
    Elias Reich\orcidID{0009-0000-4561-9067}  \and
    Saverio Messineo\orcidID{0000-0003-1592-4428} \and
    Stefan Huber\orcidID{0000-0002-8871-5814}
}

\authorrunning{E. Reich et al.}

%
\institute{
    Josef Ressel Centre for Intelligent and Secure Industrial Automation\\ 
    Salzburg University of Applied Sciences, Salzburg, Austria\\
    \email{eliassteffen.reich@fh-salzburg.ac.at}
}
\maketitle              

\begin{abstract}
Discrete automated processes in industrial and cyber-physical systems often exhibit a
repetitive structure in which successive repetitions follow a common trajectory while differing
in duration, amplitude, and fine-scale dynamics. Such \emph{approximately periodic}
behavior poses a challenge for Gaussian Processes (GP) modeling: strictly periodic models suppress
inter-repetition variability, while non-periodic models fail to capture the strong structural
regularities required for generation. In this work, we propose a stochastic generative
model for approximately periodic time series. The model is based on a GP whose
posterior is modulated by a novel kernel. Our approach decouples intra-repetition
structure from inter-repetition variability through a two-stage construction
which yields a generative distribution with a identical mean function across repetitions,
while allowing smooth variation between repetitions.
The modeling choices are supported by an implementation in which realistic synthetic
trajectories are generated from toy datasets.
\end{abstract}

\section{Introduction}

Discrete automated processes, such as those encountered in industrial production machines,
robotic assembly lines, and cyber-physical control systems, often exhibit a characteristic
form of repetition. Due to sensor noise, mechanical wear, environmental fluctuations, and
control variability, such processes are not strictly periodic. Still, their observable
dynamics consist of successive iterations that are similar in shape and duration. As a
result, measurements collected during operation give rise to multivariate time series that
repeatedly traverse comparable trajectories while exhibiting non-negligible variation
across repetitions. In this setting, \acp{gp}~\cite{rw-gp} are an appealing model: they provide
a principled probabilistic model with calibrated uncertainty and interpretable structure,
and periodic or quasi-periodic kernels allow us to encode repetition while retaining
flexibility for repetition-to-repetition variability. The task of this paper is to propose a novel,
stochastic generative model based on the \ac{gp}, that formalizes and reproduces processes
whose iterations \textquote{nearly} repeat themselves.

To the best of our knowledge, no prior GP-based generative model provides: (i)
long-horizon generative stability across arbitrarily many repetitions, and (ii) controlled,
smoothly decaying inter-repetition variability. Our kernel-modulated posterior addresses this
gap and is practically relevant in several application domains: it enables the generation
of realistic synthetic data for unit testing, digital twins or populating honeypots in
industrial systems; and it naturally supports likelihood-based anomaly detection by
quantifying deviations from learned statistics.

\paragraph{Problem formulation.}

Formally, observations of systems with repetitive behavior can be characterized as
multivariate time series over a finite interval
\begin{equation*}
  x \colon I\to\mathbb{R}^D \quad\text{with}\quad I=[0,N] \subset \mathbb{R},
\end{equation*}
that can be decomposed into consecutive segments $I_1, I_2, \dots, I_r$, with $\bigcup_j
I_j = I$ and each interval $I_j$ corresponding to an individual repetition of the underlying
process. We do not require that the $I_j$ have equal length. By introducing
reparameterizations $\pi_i \colon I_i \rightarrow [0,1]$ that normalize time, we define
the \emph{time-normalized repetition} as $ y_i(t) = x\left(\pi_i^{-1}(t)\right) $ for all $t\in
[0,1]$. Given $\varepsilon >0$, we call $x$ \emph{repetitive} and, in turn,
$y_1,\dots,y_r$ repetitions of an \emph{approximately periodic} time series, if and only
if $\| y_i - y_j \|_{\infty} \leq \varepsilon$ for all $1 \leq i, j \leq r$. In such a
case, a pair $(y_i, y_j)$ is called \emph{approximately equivalent}. The process of
splitting $x$ in its repetitions is shown in~\cite{SRMH25}. To keep the proposed method as
streamlined as possible, we will from now on assume that all time series data are already
split into their repetitions and reparametrized to satisfy the constraints for approximate
periodicity. Avoiding notational clutter, we will say $y$ is a vector which contains all
repetitions in order $[y_1,\dots,y_r]^\top$ and $T$ is the vector of their respective
domains $[T_1,\dots, T_r]^\top$. We will also assume $D=1$, but the model can be employed in
higher dimensions through multi-output GPs such as the \ac{lmc}~\cite{alvarez-k}.

Generating approximately periodic processes poses a challenge for GP models.
Classical periodic models enforce exact repetition and therefore fail to represent
realistic inter-repetition variability, while generic non-periodic models disregard the strong
structural regularities present in the data and therefore fail to make long term
predictions. From a generative modeling perspective, this is particularly limiting: the
model should be capable of producing arbitrarily long sequences of repetitions that
preserve the characteristic shape of the process, while allowing variation between repetitions.

\paragraph{Contributions.}
In this paper, we introduce a stochastic generative model for \emph{approximately
periodic} multivariate time series based on GPs, that can generate time series of
arbitrarily many repetitions, preserving continuity over the whole time span while still
allowing variation between repetitions. Our approach consists of two conceptually separate
stages: (1) learning a stable repetition template using a strictly periodic GP posterior, and
(2) injecting controlled variability between repetitions by modulating the posterior
covariance with a smooth envelope. This separation ensures that the model’s mean dynamics
remain consistent across arbitrarily long horizons, while its covariance structure
reflects the natural decay of similarity between repetitions. 
This yields a generative model that produces trajectories with identical mean structure
across repetitions while allowing smooth variation between repetitions. 

\section{Related Work}
Periodic kernels and their quasi-periodic variants are long-standing tools for modeling
repetitive structure in time series~\cite{rw-gp}. In astronomy and astrophysics, scalable
one-dimensional GP methods enable quasi-periodic modeling of stellar rotation and
activity over long light curves~\cite{foreman-17}. Recent analyses connect quasi-periodic
kernel hyperparameters to physically meaningful timescales and study their identifiability
and sampling effects~\cite{nicholson-22}. While effective and efficient algorithms for
forecasting and uncertainty estimation exist~\cite{deisenroth-14, foreman-17, li-23},
these methods either enforce exact periodicity or introduce mean decay and variance
inflation over long horizons, limiting their use as \emph{generative} models for
arbitrarily many repetitions.

A substantial line of work recasts temporal GP regression as linear state-space
models, yielding $\mathcal{O}(N)$ inference via Kalman filtering and smoothing, with
explicit linkages for periodic/quasi-periodic covariances~\cite{solin-14}. These methods
improve scalability and bring physical insight, but they do not by themselves enable
long-horizon generation.

Neural ODEs~\cite{chen-18} and diffusion/score-based models~\cite{song-20} achieve
impressive generative performance but often trade off interpretability and calibrated
uncertainty as well as long-term stability~\cite{SRUH25}. For safety-critical industrial and
cyber-physical applications, where stability is central, \acp{gp} remain attractive.

\section{Methodology}
We refer to the standard textbook~\cite{rw-gp} on GPs. Below we give a short GP
primer to introduce the notation used throughout the paper. Let training inputs be
$T=[t_1,\dots,t_n]^\top$ with observations $y=[y_1,\dots,y_N]^\top$ and define a \ac{psd}
kernel/covariance function $k(t, t')$ with Gram-matrix $K$. Evaluated on the training inputs
$K:=K(T,T)$, the matrix has entries $[K]_{ij}=k(t_i,t_j)$. For a test
input $t_*$ we write $k_{*}:=K(T,t_*)$ for the cross covariances and $k_{**}:=K(t_*,t_*)$
for the covariances. If there are multiple test inputs, we refer to them as $T_*$. With
i.i.d. Gaussian observation noise $\epsilon\sim\mathcal{N}(0,\sigma^2 I)$ and prior over
the latent (unobserved, noiseless) function $f\sim\mathcal{GP}(\mu,k)$ -- in
practice the mean function $\mu$ is usually zero without loss of generality -- the
standard posterior (i.e., predictive) distribution at $t_*$ is available in closed form:
\begin{align}\label{eq:gp_post}
  f(t_*)\mid T,y &\sim \mathcal{N}(\hat\mu,\,\hat\Sigma) \quad \text{with}\nonumber\\
  \hat\mu        &= \mu(t_*) + k_*^\top (K + \sigma^2 I)^{-1} (y - \mu(T)),\\
  \hat\Sigma     &= k_{**} - k_*^\top (K + \sigma^2 I)^{-1} k_*.\nonumber
\end{align}
Throughout this paper, we will rarely consider univariate inputs and outputs, if so it is
explicitly stated; otherwise $T=[T_1,\dots, T_r]^\top$ will denote the inputs for $r$
repetitions $y = [y_1,\dots,y_r]^\top$.\\

The core modeling challenge is to obtain a GP posterior that is approximately periodic in
the sense that
\begin{enumerate}[label=(\roman*)]
    \item the posterior mean is identical across repetitions, and\label{cond:i}
    \item the posterior covariance permits variation between repetitions.\label{cond:ii}
\end{enumerate}

A prior for periodic processes with period length $p$, kernel width $l_\theta$ and signal
variance (average distance from the prior mean) $\sigma^2_f$; is an exponential kernel
with inputs embedded on the circle~\cite{rw-gp}. We will consider $\theta=(l_\theta,
\sigma^2_f, \sigma^2_\theta)$, where $\sigma^2_\theta$ is the prior noise variance
from~\eqref{eq:gp_post}, as optimizable hyperparameters. The periodic kernel can be
written as
\begin{equation}
    k_\theta(t, t') = \sigma_f^2\exp\left(\frac{2}{l_\theta^2}\sin\left(\frac{\pi}{p}(t - t')\right)^2\right)~.
\end{equation}
Under this prior, the posterior predictive mean is exactly periodic, and both variances
and covariances repeat identically at integer multiples of the period duration $p$.
Consequently, uncertainty is propagated forward without attenuation: predictions at
arbitrarily large horizons exhibit the same uncertainty as predictions near the observed
data. This behavior is often undesirable in regression settings, where uncertainty is
typically expected to grow. For generative modeling however, this behavior is beneficial.
A generative model is expected to reproduce the empirical statistics of the data
indefinitely; enforcing increasing uncertainty over time would eventually destroy the
characteristic structure of the generated trajectories.

An additional property of the periodic kernel is that for grid-sampled data (all $r$
repetitions obtained as a sequence with equal sampling distances) the joint covariance
across the concatenated observations has a block structure: the diagonal blocks equal
$K+\sigma^2I$ (the per-repetition covariance matrix plus observation noise) and every
off-diagonal block equals the shared covariance matrix $K$. This block structure simplifies
algebra and enables the following strong statement about the predictive mean:

\begin{property}
  \label{proprerty1}
  If data is sampled on a grid then the posterior mean evaluated on the grid will converge
  to the empirical mean as the observation noise $\sigma^2\to 0$. Computing the posterior
  mean and covariance on $r$ repetitions is equivalent to computing them on one
  repetition, scaling $\sigma^2$ by $\frac1{r}$ and replacing $[y_1, \dots y_r]^\top$ by
  the mean repetition $\frac1{r}\sum_{i=1}^{r} y_i$.
\end{property}

\begin{proof}
  The proof is deferred to the appendix \cref{sec:proof-prop:mean-faithful-convergence}.
\end{proof}

Although \cref{proprerty1} was derived under the assumption of repeated observations on a
fixed grid, it provides useful intuition for more general settings. In particular, it
shows that a GP with a periodic kernel naturally encodes variability between repetitions
through the noise parameter~$\sigma^2$. Because the periodic kernel maps phase‑aligned
inputs from different repetitions to the same points in feature space, the model
effectively receives multiple (potentially different) observations at those same
locations. Any mismatch between these repeated values must then be explained by the
observation noise, and is therefore captured in~$\sigma^2$.

\begin{figure}[h]
    \centering
    \includegraphics[width=\textwidth]{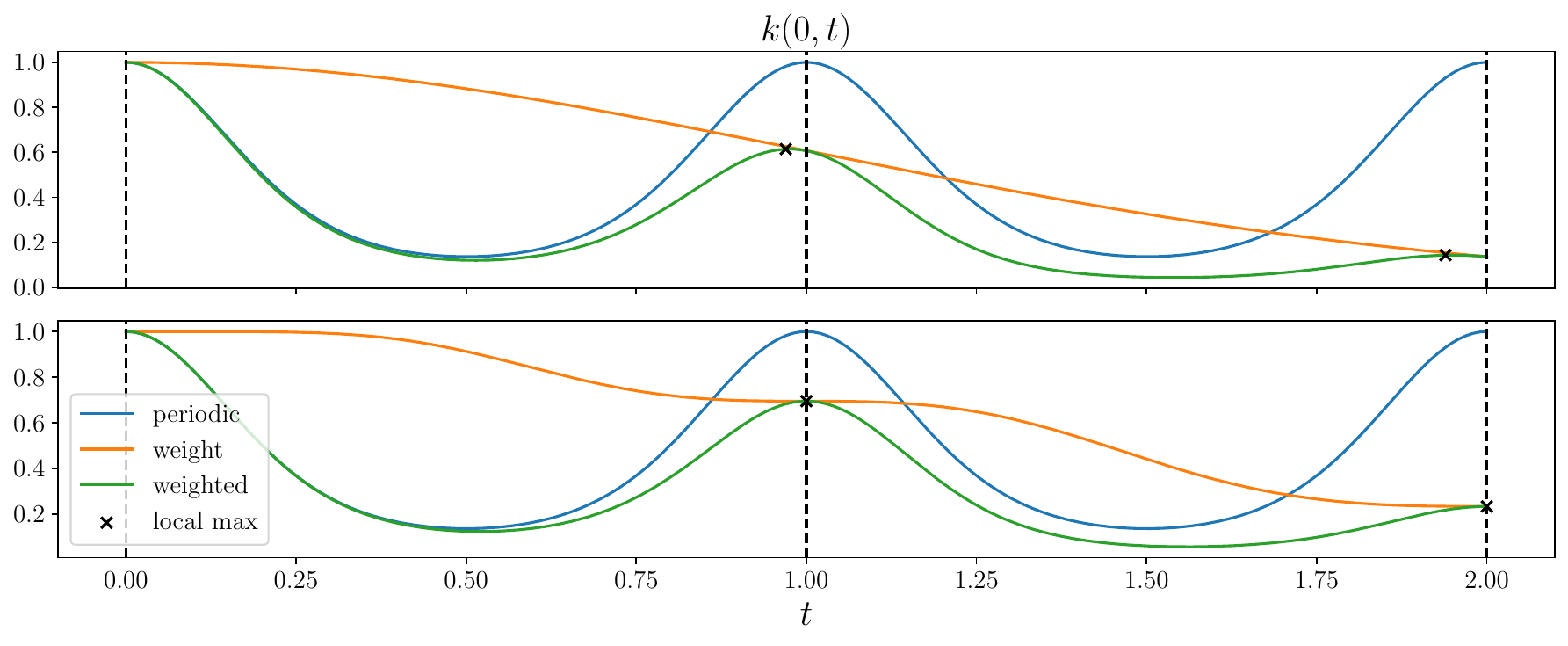}
    \caption{The top row shows the periodic kernel $k_\theta$ and squared exponential
    kernel (weight) over two periods ($p=1$) as well as the product of the two kernels
    (weighted), all centered around 0. The bottom figure replaces the standard squared
    exponential kernel with the proposed weight $w_\psi$.} \label{fig:weight_kern}
\end{figure}

To relax strict periodicity, we introduce a smoothly decaying correlation envelope $w$
with hyperparameters $\psi=(l_\psi, \sigma^2_\gamma)$ that decreases the covariance
between repetitions:
\begin{equation}
  w_\psi(t, t') = \sigma^2_\gamma\exp\left(-\frac{\| \phi(t) - \phi(t')\|^2}{2l_\psi^2}\right) \quad\text{with}\quad
    \phi(t) =\int_0^t \sin\left(\frac{\pi}{p}\tau\right)^2 ~\text{d}\tau~.
\end{equation}
The kernel $w$ is a Gaussian kernel in an embedding space, which has vanishing derivatives
at integer multiples of $p$ to avoid unwanted phase shifts in the weighted kernel
$g=w\cdot k$. Intuitively, the mapping stretches and contracts the time axis in a way that
preserves periodic structure but gradually attenuates long-range interactions,
\Cref{fig:weight_kern} illustrates the shift in local covariance maxima.

\begin{figure}[h]
    \centering
    \includegraphics[width=\textwidth]{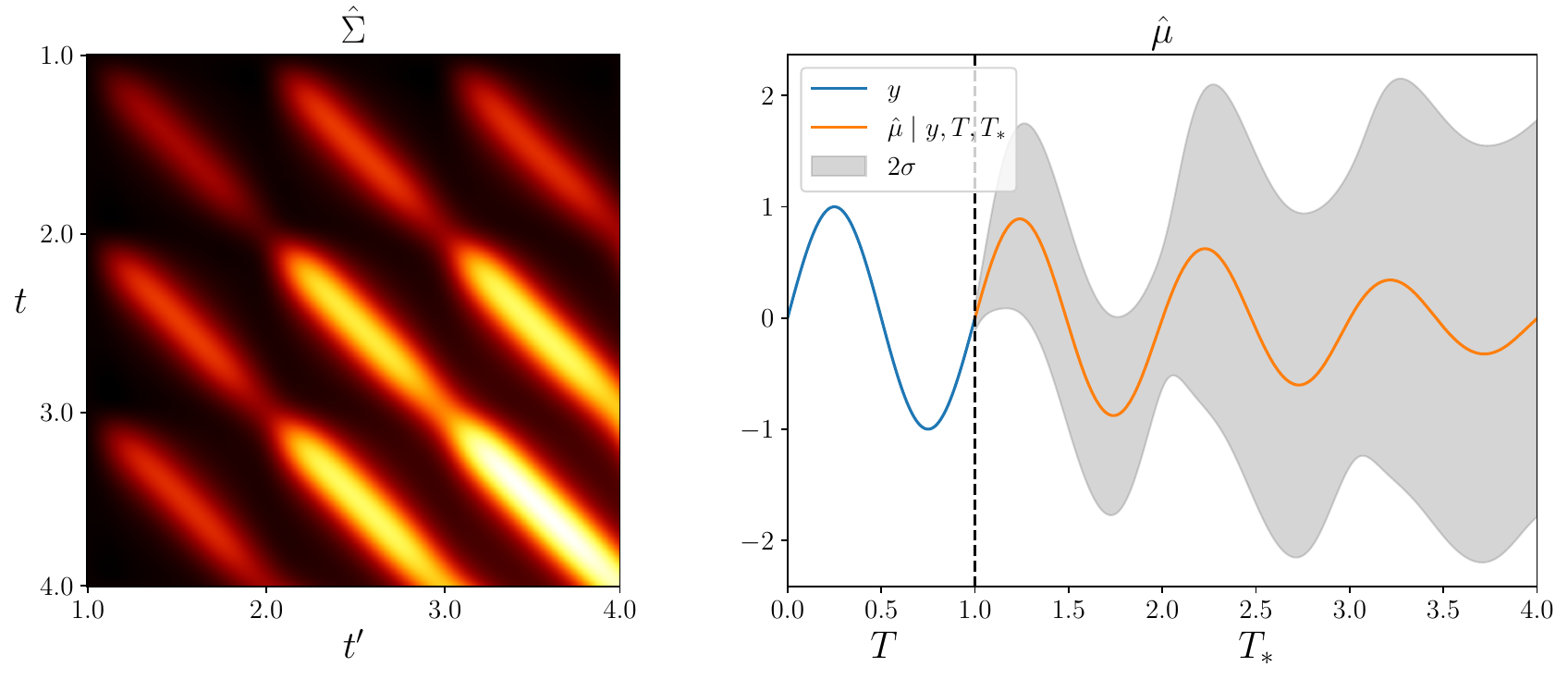}
    \caption{Posterior covariance matrix $\hat\Sigma$ and mean $\hat\mu$ with a weighted
    periodic prior covariance at test locations $T_*$. Variances as well as covariances
    are increasing and the predictive mean decays.}
    \label{fig:weighted_prior_post}
\end{figure}

Naively using $g$ as a prior will have the effect of a decaying posterior mean $\hat\mu$
and increasing posterior variances $\sigma^2=\text{diag}(\hat\Sigma)$, see \cref{fig:weighted_prior_post}. To illustrate this behavior, consider the single point
prediction at test inputs $t^*$ given observations $(T, y)$~\cite{rw-gp}:
\begin{align}
  \label{naive_use}
  \begin{split}
    p(f(t^*)|T, y) &= \mathcal{N}(\hat\mu,\hat\Sigma) \quad \text{with} \\
    \hat\mu        &= g(T, t^*)^\top g(T, T)^{-1} y = \sum_{i} g(T_i, t^*)\;g(T, T)^{-1}_{i,:} y, \\
    \hat\Sigma     &= g(t^*, t^*) - g(T, t^*)^\top g(T, T)^{-1} g(T, t^*)  \\
                   &= g(t^*, t^*) - \sum_{i,j} g(T_i, t^*)\,g(T_j, t^*)\;g(T, T)^{-1}_{i, j}
  \end{split}
\end{align}

In the mean expression in~\eqref{naive_use}, observe that the quantity computed is simply
a weighted sum of the values in the weighted periodic covariance matrix $g(T_i, t^*)$. As
the difference $|T_i-t^*|$ increases, the covariance tends to zero. The predictive mean
will thus decay to the prior (zero-) mean function, violating condition \ref{cond:i}. For
the predictive covariances the quadratic term will vanish for small $|T_i-t^*|$ and thus
tends to $g(t^*, t^*)$, i.e., the maximum value the covariance matrix can take, hence
violating condition \ref{cond:ii}. Thus, a periodic kernel with decaying envelope implies
a posterior with decaying mean and increasing variances. To avoid this pathology, we
decouple the intra- and inter-repetition statistics by adopting a two-stage construction. 

\subsection{Stage 1: Periodic prior \ac{gp}}
In the first stage, we model the intra-repetition statistics using a strictly periodic \ac{gp} prior,
\begin{equation}
    y(t) = f(t) + \epsilon,
    \qquad
    f \sim \mathcal{GP}(0, k_\theta(\cdot, \cdot)),
    \qquad
    \epsilon \sim \mathcal{N}(0,\sigma_\theta^2),
\end{equation}
and compute the posterior $f_* \mid y \sim \mathcal{N}(\hat\mu_\theta,
\hat\Sigma_\theta)$. This stage yields an exactly periodic posterior.

Even though approximately periodic time series data typically consists of a single long
trajectory, we will first interpret the observed repetitions $y_1,\dots,y_r$ as
(partially) independent realizations $f_1, \dots, f_r$ of
the same underlying periodic process,
\begin{equation}
    \forall i\in\{1,\dots,r\}\colon y_i(t) = f_i(t) + \epsilon,~t \in T_i
\end{equation}
where $f_i\sim \mathcal{GP}(0, k_\theta)$ are independent draws of the latent function
sharing common hyperparameters. However, a caveat of the fully factorized likelihood
implied by independence is that the noise variance $\sigma^2_\theta$ is not identifiable
from individual repetitions. To mitigate this issue while retaining computational
tractability, we adopt a mini-batched training strategy in which the \ac{nll} is evaluated
over small batches $\mathcal{B}$ of multiple repetitions. By jointly evaluating the kernel
on the inputs of several repetitions, the likelihood $\theta\mapsto
\mathcal{N}(y\mid\theta)$ has to explain these variations, thereby stabilizing the
estimation of the noise variance 
\begin{equation}
    \argmin_{\theta}
    \sum_{\substack{y_\mathcal{B}\subseteq y\\T_\mathcal{B}\subseteq T}} 
    -\log\mathcal{N}(y_\mathcal{B} \mid 0, K_\theta(T_\mathcal{B}, T_\mathcal{B}) + \sigma^2_\theta I).
\end{equation}

Hyperparameter optimization yields a variance estimate that captures the intrinsic
variability of the underlying process. During posterior construction, however, repetitions
must be treated as observations of the same latent function in order to preserve the
global spatial structure of the trajectory across repetitions. When using a periodic
kernel, conditioning on multiple repetitions results in an unintended accumulation of
evidence, which leads to an artificial attenuation of posterior uncertainty
(\cref{eq:post_cov_grid}). To compensate for this effect, we rescale $\sigma^2_\theta$ by
the effective number of repetitions after training. This adjustment restores uncertainty
levels consistent with those identified during hyperparameter optimization.

\subsection{Stage 2: Posterior weighting}
In the second stage, inter-repetition correlations are introduced \emph{after
conditioning} by modifying the posterior covariance. Rather than altering the GP prior or
kernel, we model these dependencies directly at the level of the posterior distribution.
Using the weight kernel $w_\psi$ only at the posterior stage preserves exact periodicity of the
mean, while avoiding pathological behavior of weighted periodic priors, which cause mean
decay and variance inflation. This choice is essential for maintaining long-horizon
stability. Specifically, we define the following generative model:
\begin{equation}
    \gamma \sim \mathcal{N}(\hat\mu_\theta,\, \Sigma_\psi:= W_\psi \odot \hat\Sigma_\theta + \sigma^2_\psi I),
\end{equation}
where $W$ is the kernel matrix induced by $w_\psi$, and $\odot$ denotes elementwise
multiplication (Schur product). We do not claim this construction defines a GP,
i.e., no GP prior with this posterior exists. By construction, $\gamma$ is a random
variable, specified at a finite set of evaluation points and follows a non-degenerate
normal distribution, as the modified posterior covariance matrix is \ac{psd} by the Schur
product theorem. 

By leaving the posterior mean unchanged, the model enforces identical mean structure
across all repetitions. By weighting the covariance, local uncertainty is preserved while
ensuring that correlations between distant repetitions decay monotonically to zero. This
behavior matches the statistical structure expected of approximately periodic data and
avoids the degeneracies induced by envelope-weighted GP priors. An alternative
construction, the additive combination of $W$ and $\hat\Sigma$, is simpler from a modeling
perspective but fails to eliminate long-range periodic correlations: once the envelope
decays, the covariance matrix reverts to that of a strictly periodic GP posterior. The
proposed multiplicative covariance modulation is therefore essential for suppressing
long-range periodic correlations while preserving phase-aligned variance, resulting in a
generative model whose second-order statistics more faithfully reflect those observed in
approximately periodic data.

The strength of long range correlation dampening is decided by the length scale
hyperparameter $l_\psi$ of the weighting kernel and independent noise in the outputs is
learned via the hyperparameter $\sigma^2_\psi.$\footnote{This parameter also prevents a
degenerate solution for the dampening ($l_\psi\to 0$) in a noiseless setting. We recommend
to discard $\sigma^2_\psi$ after training if the generated trajectories need to be without
noise.}
As in the single repetition model, we find their optima by \ac{nll} minimization. In
contrast to the first stage, the observations $y_1,\dots,y_r$ are now no longer treated as
independent, as this would preclude inference on the inter-repetition correlations.
\begin{equation}
    \argmin_{\psi}~(y-\hat\mu_\theta) \Sigma_\psi^{-1}(y-\hat\mu_\theta)^\top+\log\det\Sigma_\psi
\end{equation}
We observer that the inverse and determinant term involving the Schur product ($\Sigma_\psi$) admit no
algebraic simplification. Evaluating the likelihood is thereby problematic with a large
number of repetitions.

\section{Experimental Setup}
We evaluate the influence of the first-stage mini-batch size and the amount of output
noise that the second stage must absorb. The dataset is a trajectory
$y_\text{test}(t_*),\, t_*\in T_*$ sampled from a weighted-periodic GP conditioned on the
signal $\sin(2\pi t)$. From each of the 10 repetitions we pick 20 random points for
training. Each
stage is optimized via the two-stage \ac{nll} procedure; learned hyperparameters and
\ac{mse} between predictive and test mean $\mu_\text{test}, \mu_\theta$ and pointwise
standard deviations $\hat\sigma_\text{test}:=\sqrt{\text{diag}(\Sigma_\text{test})},
\hat\sigma_\gamma:=\sqrt{\text{diag}(\Sigma_\psi)}$ are reported for evaluation. The models are
optimized using Adam (lr $= 0.1$) for 100 gradient steps in each
stage\footnote{code available on \url{https://github.com/JRC-ISIA/2026-lion20-approx-periodic-gp-generative-model}}.

\begin{figure}[h]\label{fig:test_data}
    \centering
    \includegraphics[width=.95\textwidth,trim={3mm 3mm 2mm 2.5mm},clip]{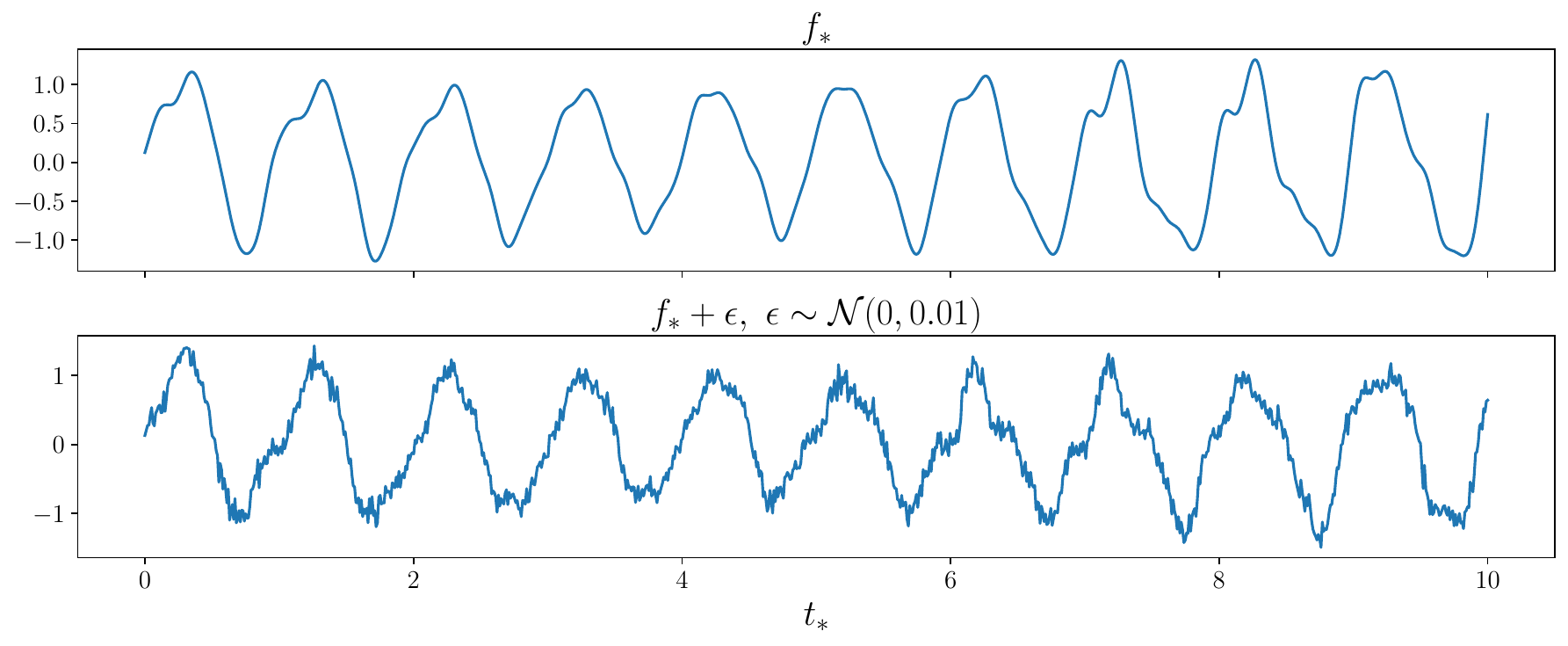}
    \caption{Approximately periodic test data without noise (top) and with noise (bottom).}
 \end{figure}
For noiseless approximately periodic data, we set the hyperparameters to $l_k=0.7,
l_w=0.8, \sigma^2=1.5$
\begin{align}
    y(t) &= f(t) + \epsilon,\quad f\sim\mathcal{GP}(0, k(t, t')w(t, t')),\quad\epsilon\sim\mathcal{N}(0, \sigma^2) \nonumber\\
    y_\text{test}(t_*) &= f_*(t_*), \quad f_*\sim\mathcal{N}(\mu_\text{test}, \Sigma_\text{test}).
\end{align}
For the second experiment, we add noise $\sigma_\text{out}^2$ to the test data and fix the
batch size $|\mathcal{B}| = 2$, as we see diminishing returns for larger numbers.
\begin{align}
    y_\text{test}(t_*) = f_*(t_*) + \epsilon_\text{out},\quad\epsilon\sim\mathcal{N}(0, \sigma_\text{out}^2)~.
\end{align}
We also include results for periodic data, where we use the same hyperparameters and drop
the weight kernel in the prior.

The final experiment follows a similar construction as the previous ones, but with
2-dimensional data. We condition on $[\sin(2\pi t), -\sin(4\pi t)]$ and fit a \ac{lmc}
with non-stationary periodic kernel~\cite{rw-gp, paciorek-03}. The coregionalization
matrix is optimized in the first stage. For details on multi-output \acp{gp}
see~\cite{alvarez-k}.

\begin{table}[t]
    \centering
    \caption{Results on approximately periodic data with varying batch size $|\mathcal{B}|$.}
    \label{tab:approx_periodic}
    \begin{tabular}{lrrrrrrrr}
        \toprule
        $|\mathcal{B}|$ &\quad
        MSE($\mu_\text{test}, \mu_\gamma$) &\quad
        MSE($\hat\sigma_\text{test}, \hat\sigma\gamma$) &\qquad
        $l_\theta$ &\qquad
        $l_\psi$ &\qquad
        $\sigma^2_\theta$ &\qquad
        $\sigma^2_\psi$ &\qquad
        $\sigma^2_f$ &\qquad
        $\sigma^2_\gamma$\\
        \midrule

        1 & \num{1.88e-02} & \num{2.45e-02} & 0.65 & 0.609 & 0.009 & 0.027 & 0.504 & 8.05\\
        2 & \num{1.34e-02} & \num{4.33e-04} & 0.982 & 0.687 & 0.307 & 0.0 & 0.587 & 5.559 \\
        3 & \num{1.36e-02} & \num{1.04e-03} & 0.937 & 0.69 & 0.383 & 0.0 & 0.663 & 4.946 \\
        5 & \num{1.27e-02} & \num{2.13e-03} & 1.037 & 0.694 & 0.423 & 0.0 & 0.639 & 5.467\\
        10 & \num{1.31e-02} & \num{1.93e-03} & 0.891 & 0.695 & 0.463 & 0.0 & 0.513 & 4.69\\
        \bottomrule
    \end{tabular}
\end{table}

\begin{table}[t]
    \centering
    \caption{Results on approximately periodic data with noise $\sigma^2_\text{out}$ and $|\mathcal{B}|=2$.}
    \label{tab:approx_periodic_noise}
    \begin{tabular}{lrrrrrrrrr}
        \toprule
        $\sigma^2_\text{out}$ &\quad
        MSE($\mu_\text{test}, \mu_\gamma$) &\quad
        MSE($\hat\sigma_\text{test}, \hat\sigma\gamma$) &\qquad
        $l_\theta$  &\qquad
        $l_\psi$  &\qquad
        $\sigma^2_\theta$ &\qquad
        $\sigma^2_\psi$ &\qquad
        $\sigma^2_f$ &\qquad
        $\sigma^2_\gamma$
        \\
        \midrule

        0.01 & \num{1.35e-02} & \num{3.85e-04} & 0.945 & 0.677 & 0.304 & 0.0 & 0.56 & 5.411 \\
        0.05 & \num{1.48e-02} & \num{3.97e-04} & 0.895 & 0.634 & 0.34 & 0.003 & 0.611 & 4.726 \\
        0.10 & \num{9.63e-03} & \num{4.87e-04} & 1.027 & 0.67 & 0.396 & 0.01 & 0.706 & 4.617 \\
        0.30 & \num{1.11e-02} & \num{9.06e-04} & 1.544 & 0.587 & 1.117 & 0.088 & 0.839 & 1.2 \\
        0.50 & \num{2.53e-02} & \num{1.78e-03} & 1.184 & 0.098 & 3.268 & 0.222 & 0.674 & 1.936 \\
        \bottomrule
    \end{tabular}
\end{table}

\begin{table}[t]
    \centering
    \caption{Hyperparameters via minimizing the \ac{nll} on periodic data.}
    \label{tab:periodic}
    \begin{tabular}{lrrrrp{1.5em}lrrrrr}
        \toprule
        $|\mathcal{B}|$~ &
        $l_\theta$ &\qquad
        $l_\psi$ &\qquad
        $\sigma^2_\theta$ &\qquad
        $\sigma^2_\psi$ &\qquad
        &
        $\sigma_\text{out}^2$ &\qquad
        $l_\theta$  &\qquad
        $l_\psi$  &\qquad
        $\sigma^2_\theta$ &\qquad
        $\sigma^2_\psi$ \\

        \cline{1-5} \cline{7-11}

        1   &0.922&	0.959&	0.003&	0.0& &
        0.01 & 0.801&	0.902&	0.002&	0.0  \\
        2	&0.816&	0.917&	0.001&	0.0& &
        0.05 & 1.082&	1.405&	0.033&	0.003  \\
        3	&0.819&	0.913&	0.001&	0.0& &
        0.10 & 1.909&	1.021&	0.106&	0.01  \\
        5	&0.758&	0.907&	0.001&	0.0& &
        0.30 & 2.247&	0.824&	0.902&	0.084  \\
        10	&0.679&	0.906&	0.001&	0.0& &
        0.50 & 1.623&	2.081&	2.726&	0.27  \\

        \bottomrule
    \end{tabular}
\end{table}

\subsection{Discussion}
\begin{figure}[t]
    \centering
    \includegraphics[width=\textwidth]{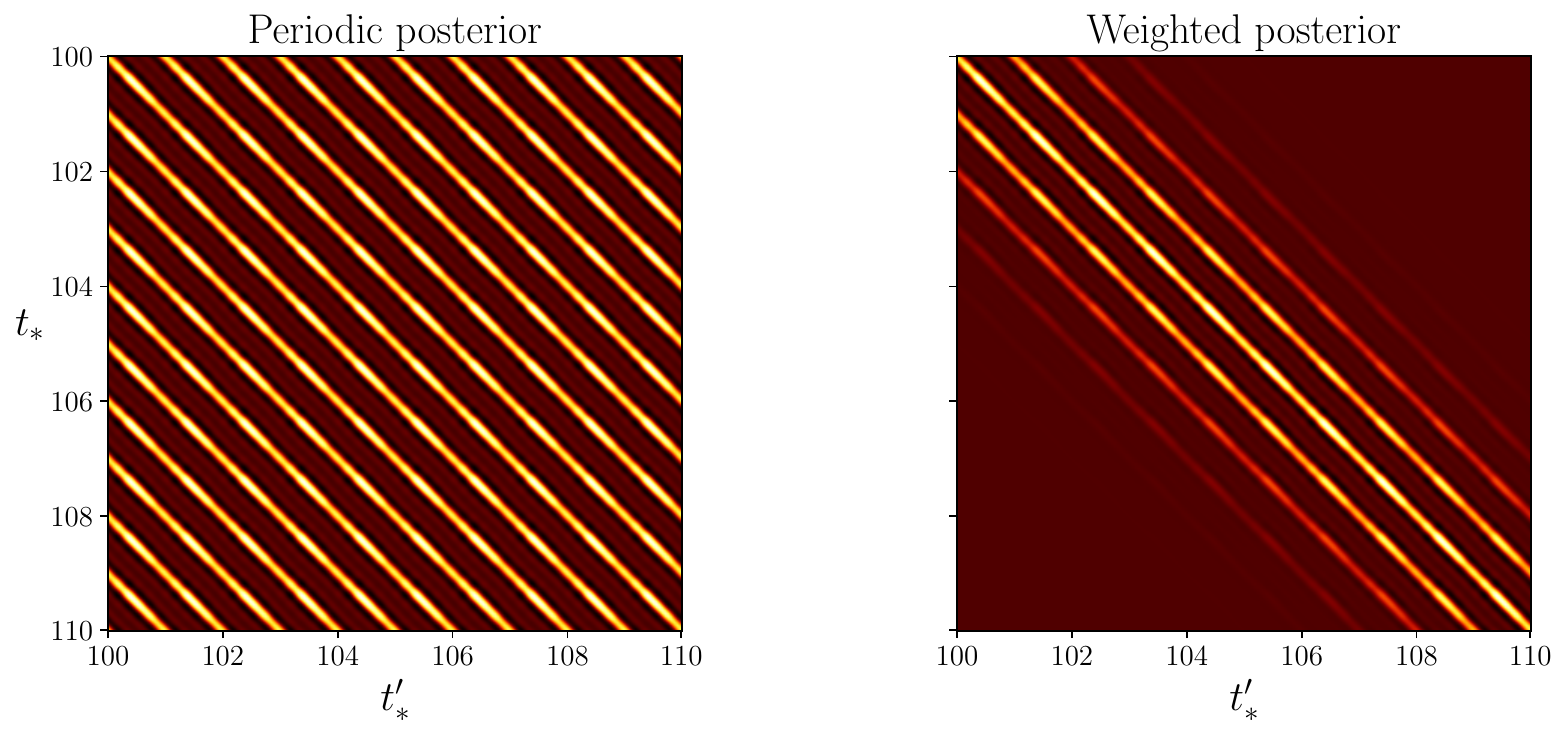}
    \includegraphics[width=\textwidth]{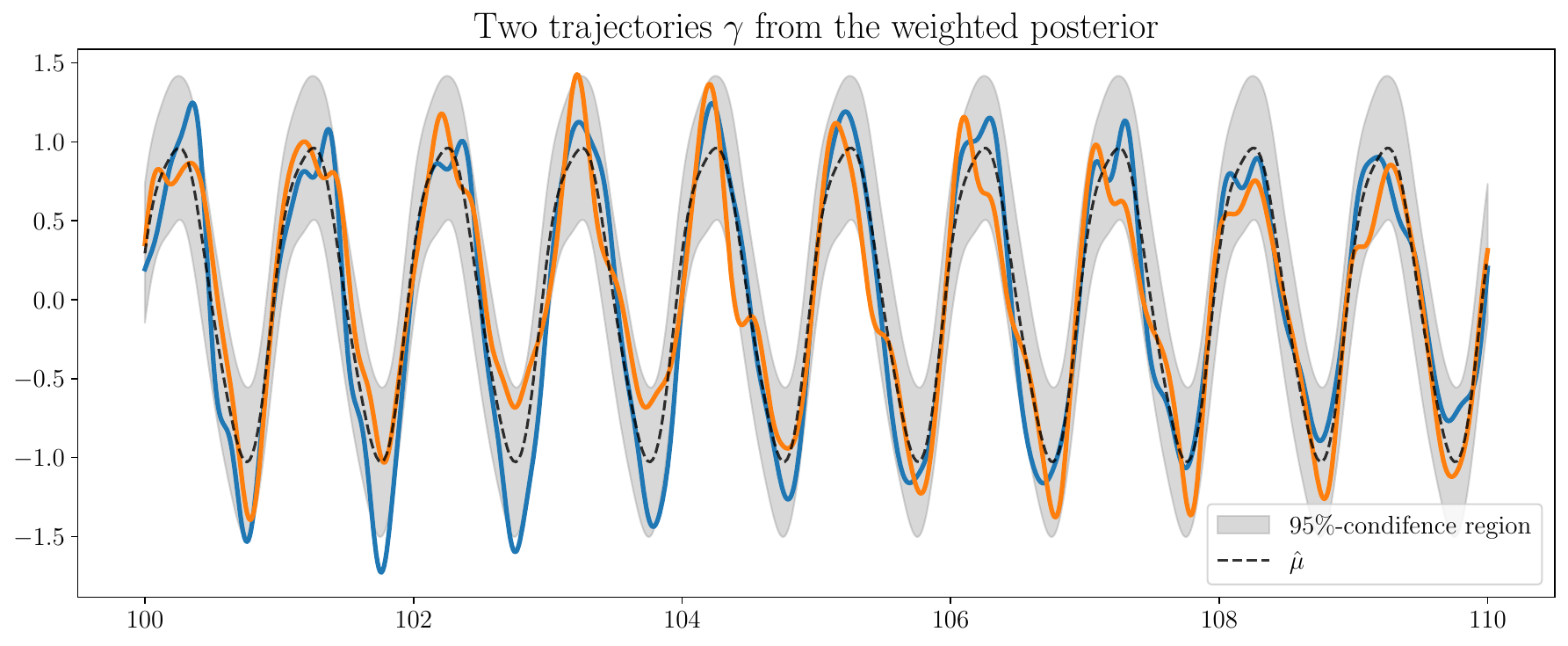}
    \caption{Posterior covariance matrix and samples from $\gamma$ with hyperparameters
    from the noiseless setting in \cref{tab:approx_periodic} with $|\mathcal{B}|=2$.}
    \label{fig:gamma_example}
\end{figure}

The results in \cref{tab:approx_periodic} demonstrate a strong dependence of the
first-stage variance estimates $\sigma_\theta^2$ on the mini-batch size $|\mathcal{B}|$.
When training with batch size $|\mathcal{B}| = 1$, the estimated variance collapses
towards zero, despite the fact that the observed repetitions exhibit substantial
inter-repetition variability. As the batch size increases, the estimated variance
stabilizes and approaches a non-degenerate value. This behavior reflects the fact that,
under a strictly factorized likelihood, each repetitions are explained by independent
latent functions. In this regime, the optimizer can reduce the \ac{nll} by ignoring
variation, leading to systematic underestimation of $\sigma^2_\theta$. Evaluating the
likelihood over mini-batches of multiple repetitions exposes the likelihood to
inter-repetition variability, providing a stronger and more stable signal for estimating
$\sigma_\theta^2$ which results in better estimates for true pointwise variances
$\hat\sigma_\text{test}^2$. The mean function $\mu_\text{test}$ is correctly identified
across all batch sizes and the learned hyperparameters are relatively consistent across
all batch sizes $|\mathcal{B}|>1$. See \cref{fig:gamma_example} for a visual of the
posterior with $|\mathcal{B}|=2$.

The second experiment in \cref{tab:approx_periodic_noise} illustrates the role of the
posterior-weighting stage under noisy observations. Whereas the the additional diagonal
variance $\sigma_\psi^2$ in the weighting kernel was consistently driven to 0 in the
noiseless setting, here its value is increasing with the additional noise put on the
observations. For small noise levels we see that the first stage observation noise
$\sigma^2_\theta$ is barely changing, as the overall pointwise variation in the data is as
high or higher than the perturbations. This behavior is then compensated by $\sigma^2_\psi$,
and only for higher noise levels ($>0.1$), the first stage observation noise is
increasing. This also explains why the signal variance $\sigma^2_\psi$ is much smaller for
these higher noise levels, as the parameter does not need to compensate the small
variances from the first stage.

For comparison, \cref{tab:periodic} (left half) reports results for strictly periodic
data. In this setting, the batch size has no influence, which is to be expected as all
repetitions are exactly equal. The posterior dampening parameter $l_\psi$ is higher as in
the first example, but we have also seen much lower values in other experiments.
Theoretically the optimum would be $\infty$, but with very low first stage prior variances
$\sigma^2_\theta$, the posterior weighting has little to no effect, as the variation for
trajectories $\gamma$ is extremely low anyways. In the noisy setting \cref{tab:periodic},
the prior observation noise $\sigma^2_\theta$ is a much better estimator for the true
noise levels, as it is not responsible for also capturing the inter-repetition variation
as well. The second stage noise variance $\sigma^2_\psi$ is also reliably increasing with
the noise level in the observations and all other hyperparameters are relatively consistent again.

\begin{figure}[t]
    \centering
    \includegraphics[width=\textwidth]{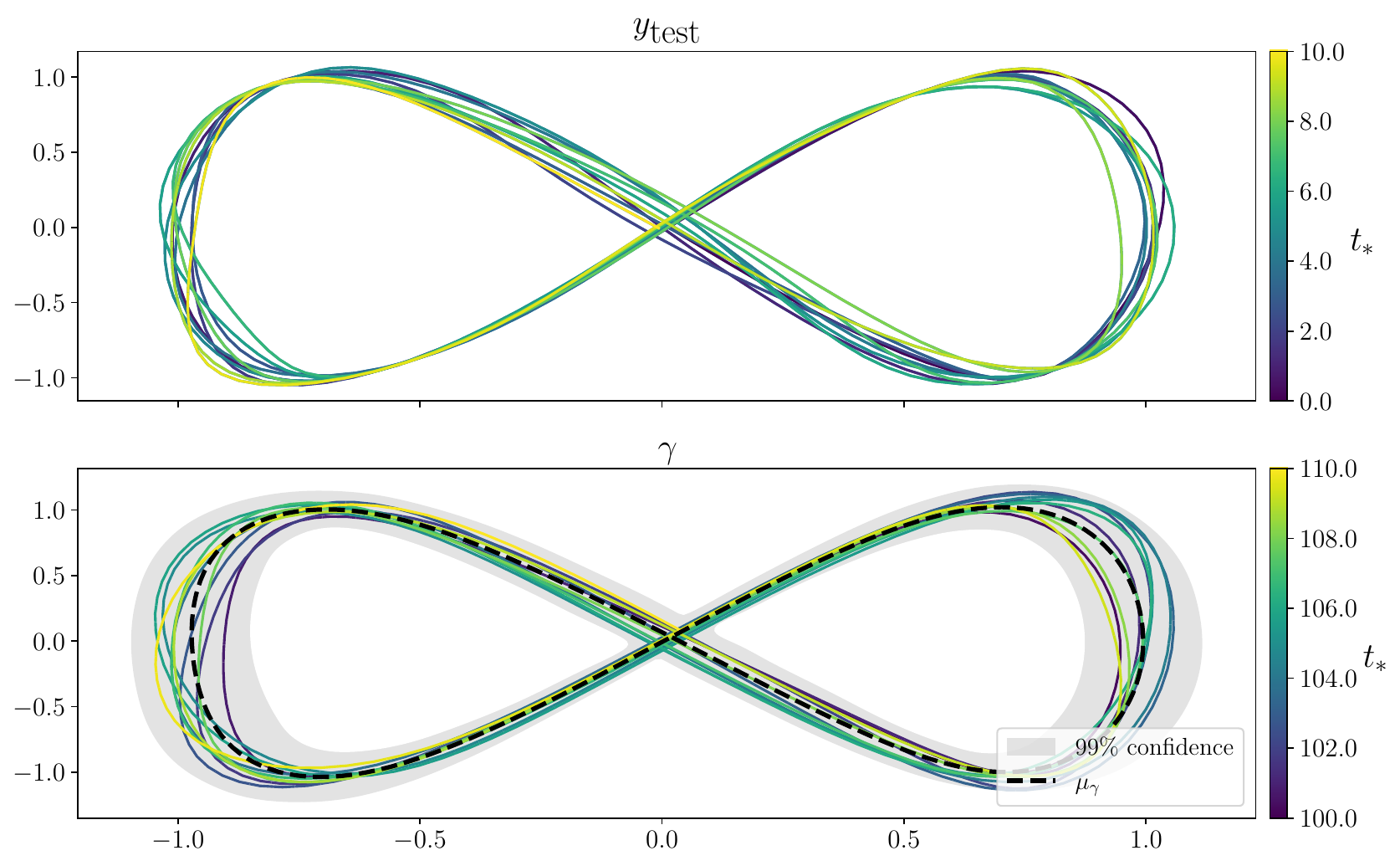}
    \caption{Test data (top) and samples from $\gamma$ (bottom). The non-stationary kernel
    can capture the high variance when crossing the origin from the second to the fourth
    quadrant.}\label{fig:2d_example}
\end{figure}

The final experiment \cref{fig:2d_example} on a 2-dimensional dataset shows how our method is applicable to more
complex problems than univariate, stationary time series. A periodic non stationary kernel
in the first stage can find the varying degrees of variation in each cycle and project
them forward indefinitely. This enables much more realistic trajectories, as naturally,
the degree of variation is not the same everywhere. The second stage model decorrelation 
then introduces variation to the otherwise periodic trajectory.

Overall, the provided experiments highlight that mini-batch likelihood evaluation plays a crucial
role in stabilizing variance estimation while keeping computational costs low in the first
stage; and that the posterior weighting mechanism in the second stage provides a
principled and effective means of modeling inter-repetition variability and output noise.

\section{Conclusion}

This paper demonstrates that modifying the GP posterior, rather than the prior,
provides a principled way to model approximately periodic processes for generative
purposes. The proposed model finds intra-repetition variability by assuming that the data
are partially independent observations of a periodic GP prior. To enable the model to generate
trajectories with variability, the posterior covariance of the periodic GP is
modulated by a novel kernel which decorrelates repetitions, breaking the exact periodicity
induced by the prior kernel. The resulting distribution is not a GP in the sense that
it is the result of conditioning a GP prior, but a non-degenerate multivariate
normal distribution derived from a GP posterior, from which trajectories of arbitrary
length may be sampled. From an application perspective, the proposed model is well suited
for synthetic data generation of stationary processes which require stability and
interpretability over long time spans.

Several directions for future work remain open. Scaling the likelihood optimization to
large numbers of repetitions will require sparse or approximate inference
techniques~\cite{titsias-09,hensman-13}, which could be adapted to the proposed covariance
modulation framework. Extending the model to handle more complex repetition statistics
and repetitive behavior could further improve expressiveness, which is inherently lacking
in the proposed model, as is the case for all stochastic processes with unimodal
posteriors. Finally, a systematic evaluation on real-world industrial datasets will be
essential to quantify the practical benefits of the approach in downstream tasks such as
anomaly detection and system simulation.


\bibliography{main}
\bibliographystyle{splncs04}


\section*{Appendix}

\subsection{Proof of \Cref{proprerty1}}
\label{sec:proof-prop:mean-faithful-convergence}

\begin{proof}
        We consider repeated noisy observations of a single latent function
        $f\sim\mathcal{GP}(0, k)$ observed on a fixed grid of size $n$. With a periodic
        kernel, the embedding on the circle will provide this grid automatically, assuming
        all repetitions are phase alinged sequences of samples at equidistant times
        covering one period length $p$ of the kernel. For repetitions $i=1,\dots, r$
        \begin{equation*}
            y_i = f + \epsilon_i,\quad \epsilon_i\sim\mathcal{N}(0, \sigma^2I)
        \end{equation*}

        The joint density of all repetitions is a block matrix with the observation noise
        $\sigma^2$ added to its diagonal.
        \begin{align*}
        [y_1, \dots, y_r]^\top &\sim \mathcal{N}\left(\bm{0}, \Sigma_r\right)\\
        \Sigma_r &= \begin{bmatrix}
            K+\sigma^2I_n & K             & \cdots & K \\
            K             & K+\sigma^2I_n & \cdots & K  \\
            \vdots        & \vdots        & \ddots & \vdots \\
            K             & K             & \cdots & K+\sigma^2I_n
        \end{bmatrix}
        \end{align*}

\begin{align*}
        \intertext{With an $r$-dimensional vector $\mathbbm{1}_r$ of all ones and the Kronecker
        product $\otimes$, we can rewrite the matrix as}
        &= (\mathbbm{1}_r\mathbbm{1}_r^\top) \otimes K + \sigma^2I_{nr} . \\
        &= \underbrace{(\mathbbm{1}_r\otimes I_n)}_U
           \underbrace{K}_C
           \underbrace{(\mathbbm{1}_r\otimes I_n)^\top}_{U^\top} +
           \underbrace{\sigma^2I_{nr}}_A~. \\
        \intertext{Recall the Woodbury identity}
        (UCV + A)^{-1} &= A^{-1} - A^{-1}U(C^{-1}+V A^{-1}U)^{-1}V A^{-1} ~,\\
        \intertext{which we apply as}
        (UCU^\top + A)^{-1} &= A^{-1} - A^{-1}U(C^{-1}+U^\top A^{-1}U)^{-1}U^\top A^{-1} ~.\\
        \intertext{so we find the inverse of $\Sigma_r$ by applying this identity twice:}
        \Sigma_r^{-1} &= \sigma^{-2}I_{nr} - \sigma^{-4} U (I_n K^{-1}I_n + \frac{r}{\sigma^2}I_n)^{-1} U^\top~.  \\
        \Sigma_r^{-1} &= \sigma^{-2}I_{nr} - \sigma^{-4} (\mathbbm{1}_r\mathbbm{1}_r^\top) \otimes
        (
        \underbrace{I_n}_U
        \underbrace{K^{-1}}_C
        \underbrace{I_n}_{U^\top} +
        \underbrace{\frac{r}{\sigma^2}I_n}_A)^{-1} ~   \\
        &= \sigma^{-2}I_{nr} - (\mathbbm{1}_r\mathbbm{1}_r^\top) \otimes
        \underbrace{\left(\frac{\sigma^{-2}}{r}I_n - \frac1{r^2} (K + \frac{\sigma^2}{r}I_n)^{-1}\right)}_{\Lambda}.
    \end{align*}

    The posterior mean on the grid then is
    \begin{align}
        \hat\mu &= (\mathbbm{1}_r^\top\otimes K) \Sigma_r^{-1}
        \begin{bmatrix}
        y_1\nonumber\\
        \vdots\nonumber\\
        y_r
        \end{bmatrix} =
        \begin{bmatrix}
            K \cdots K
        \end{bmatrix}
        \begin{bmatrix}
        \sigma^{-2} y_1 - \sum_{j=1}^r \Lambda y_j  \nonumber\\
        \vdots\nonumber\\
        \sigma^{-2} y_r - \sum_{j=1}^r \Lambda y_j   \nonumber\\
        \end{bmatrix}\nonumber\\
        &= \sum_{i=1}^r \sigma^{-2}K y_i - r\sum_{j=1}^r K\Lambda y_j =
        K\left( \sigma^{-2}I_n - r\Lambda\right)\sum_{i=1}^r y_i \nonumber\\
        &= K\left(K+\frac{\sigma^2}{r} I_n\right)^{-1} \frac1{r}\sum_{i=1}^r y_i~.
    \end{align}
    Conditioning a GP on multiple noisy observations of the same underlying latent
    function is equivalent to conditioning on the mean observation and reducing the
    observation uncertainty $\sigma^2$ by a factor of the number of observations
    $\frac1{r}$. Without observation uncertainty ($\sigma\to 0$) or in the limit of
    infinitely many observations ($r\to\infty$), the posterior predictive mean converges
    to the expected observation. With the same approach, one can also find the posterior
    covariance matrix which also simplifies nicely, leaving one with the same scaled
    variance as in the expression for the mean
    \begin{equation}
        \hat\Sigma = K - K \left(K + \frac{\sigma^2}{r}I_n\right)^{-1} K~.\label{eq:post_cov_grid}
    \end{equation}
    This reduction of effective observation variance explains why increasing the number of
    repetitions sharpens the posterior and decreases predictive variability.
\end{proof}

\end{document}